\documentclass[conference]{IEEEtran}

\usepackage[T1]{fontenc}
\usepackage{amsmath,amssymb}
\usepackage{booktabs}
\usepackage{graphicx}
\usepackage{multirow}
\usepackage{url}
\usepackage{hyperref}
\usepackage{array}
\usepackage[caption=false]{subfig}
\usepackage{flushend}
\usepackage{stfloats}

\graphicspath{{img/}}

\title{BabyMamba-HAR: Lightweight Selective State Space Models for Efficient Human Activity Recognition on Resource Constrained Devices}

\author{
  \IEEEauthorblockN{Mridankan Mandal}
  \IEEEauthorblockA{Department of Information Technology\\
  Indian Institute of Information Technology, Allahabad\\
  Prayagraj, India\\
  mridankanmandal2006@gmail.com}
}

\begin{document}
\maketitle

\begin{abstract}
Human activity recognition (HAR) on resource constrained devices requires high accuracy across diverse sensor setups. Selective state space models (SSMs) offer efficient linear time sequence processing, presenting a compelling alternative to attention mechanisms. However, their TinyML design space remains unexplored. This paper introduces BabyMamba-HAR, comprising two lightweight architectures: (1) CI-BabyMamba-HAR, utilizing a channel independent stem for noise robustness, and (2) Crossover-BiDir-BabyMamba-HAR, utilizing an early fusion stem for channel count independent complexity. Both integrate weight tied bidirectional scanning and gated temporal attention pooling. Across eight benchmarks, Crossover-BiDir-BabyMamba-HAR averages an 86.52\% F1-score with 27K parameters and 2.21M MACs, matching TinyHAR (86.16\%) while requiring 11$\times$ fewer MACs on high channel datasets. On-device deployment on the Raspberry Pi Pico~2 and ESP32 utilized a mixed precision C++ runtime (INT8 projections, float32 states). A fused computation strategy with lifetime aware memory management reduces peak memory footprint from $\mathcal{O}(B \cdot d_{\text{model}} \cdot L \cdot d_{\text{state}})$ to $\mathcal{O}(B \cdot d_{\text{model}} \cdot d_{\text{state}})$, adapting to support weight-tied bidirectional and channel-streaming execution. Both architectures achieved full 8/8 dataset coverage with $>$99.2\% PyTorch parity, whereas INT8 quantized TFLite baselines showed degraded coverage and parity (TinyHAR: 7/8 and 4/8 coverage at 60.4\% and 88.6\% parity, TinierHAR: 8/8 and 6/8 at 54.2\% and 90.8\%, DeepConvLSTM: 1/8 and 0/8 on Pico~2 and ESP32, respectively). Crossover-BiDir-BabyMamba-HAR averages 154.4~ms latency on ESP32 and 481.9~ms on Pico~2. Ablations confirm bidirectional scanning and gated attention improve F1-scores by up to 8.42\% and 8.94\%, respectively, establishing practical principles for TinyML SSM deployment.
\end{abstract}


\section{Introduction}

Human activity recognition (HAR) constitutes a foundational capability for mobile health monitoring, assistive technologies, and industrial safety systems, where motion patterns must be classified from multi-channel inertial sensor streams in real time. The deployment of HAR models on resource constrained wearable devices is strongly motivated by privacy considerations, raw sensor data can reveal sensitive behavioral patterns, and by latency requirements in safety critical applications. However, battery powered embedded systems impose severe constraints on model size (typically $<$100KB Flash) and computational cost (typically $<$10M MACs per inference), which has catalyzed the development of TinyML oriented HAR architectures \cite{zhou2022tinyhar,tinierhar2025,muhoza2023power}.

A persistent challenge in this domain is maintaining both efficiency and accuracy across heterogeneous benchmarks, where channel counts range from 3 (smartphone accelerometer) to 79 (full body sensor network), sampling rates vary from 20~Hz to 100~Hz, and sensor placement introduces varying correlation and noise structures. Furthermore, inconsistent evaluation protocols across prior work have obscured true model trade-offs \cite{geissler2024beyond}, necessitating unified comparative studies.

Selective state space models (SSMs) have emerged as a promising paradigm for efficient sequence modeling. The Mamba architecture \cite{gu2023mamba} introduces input dependent discretization that enables selective state updates with linear time complexity $O(N)$, contrasted with the $O(N^2)$ complexity of self attention mechanisms \cite{vaswani2017attention}. This property aligns naturally with HAR, where discriminative information is often localized in brief motion transitions while extended stationary periods contribute primarily noise.

Despite this alignment, the design space for lightweight SSMs in HAR remains insufficiently characterized. Critical questions include: (1) How should multi-channel sensor streams be projected into the state space backbone? (2) How should bidirectionality be implemented for windowed classification without doubling parameters? (3) Which temporal aggregation strategies remain robust under extreme parameter constraints?

These questions are addressed through a systematic study of two novel BabyMamba-HAR architectures evaluated under a unified protocol across eight public datasets. The contributions of this work are:

\begin{itemize}
\item Two complementary lightweight SSM architectures: CI-BabyMamba-HAR with channel independent processing for noise robustness, and Crossover-BiDir-BabyMamba-HAR with early fusion for computational efficiency on high channel datasets.
\item A weight tied bidirectional scanning mechanism that doubles the effective receptive field without additional parameters.
\item A lightweight context gated temporal attention pooling head that focuses on discriminative timesteps.
\item Evaluation across eight benchmarks with unified preprocessing, subject independent splits, and five random seed statistical reporting.
\item Systematic ablation studies isolating the contribution of bidirectionality, pooling strategy, stem architecture, and model hyperparameters.
\item A novel embedded deployment implementation featuring a fused streaming selective scan engine with lifetime aware memory management. This runtime reduces the peak memory bottleneck from $\mathcal{O}(B \cdot d_{\text{model}} \cdot L \cdot d_{\text{state}})$ to $\mathcal{O}(B \cdot d_{\text{model}} \cdot d_{\text{state}})$, natively supporting weight-tied bidirectional and channel-streaming execution. Evaluated on the Raspberry Pi Pico~2 and ESP32 with mixed precision, this approach achieved full dataset coverage and high numerical parity, overcoming the severe deployability limitations observed in standard TFLite micro baselines.
\end{itemize}

\section{Related Work}

\subsection{Efficient HAR Architectures}

Lightweight HAR has been extensively studied through CNN and RNN compression techniques. DeepConvLSTM \cite{ordonez2016deep} established the CNN-LSTM paradigm, achieving strong accuracy, but requiring $>$130K parameters. TinyHAR \cite{zhou2022tinyhar} introduced compact 2D convolutions with temporal attention pooling, reducing parameters to $\sim$55K while maintaining competitive accuracy. TinierHAR \cite{tinierhar2025} further compressed models to $\sim$33K parameters using depthwise separable convolutions and bidirectional GRU. Additional efficiency oriented designs include knowledge distillation approaches \cite{deng2023lhar}, neural architecture search for mobile deployment \cite{lim2023efficient}, and MLP only backbones \cite{zhou2024mlphar}.

Transformer based HAR models have also been investigated, but their quadratic complexity with sequence length raises concerns for resource constrained deployment \cite{lattanzi2024transformers}. Recent work on cross dataset generalization \cite{hong2024crosshar} and evaluation protocol standardization \cite{geissler2024beyond} has highlighted the importance of unified benchmarking.

\subsection{State Space Models and Mamba}

State space models provide a principled framework for sequence modeling through structured linear recurrences. The S4 architecture \cite{gu2022s4} demonstrated efficient long range dependency modeling through HiPPO initialized state matrices. Mamba \cite{gu2023mamba} introduced selective state spaces where discretization parameters become input dependent, enabling content-aware gating with hardware efficient parallel scans.

Recent work has begun adapting Mamba to sensor based HAR. In HARMamba \cite{harmamba2024}, a bidirectional Mamba architecture for wearable sensors was proposed. However, existing SSM based HAR designs have not systematically characterized the lightweight regime ($<$30K parameters) nor provided controlled ablations isolating architectural contributions.

\section{Methodology}

\subsection{Problem Formulation}

Consider a multi-channel inertial sensor window $\mathbf{X} \in \mathbb{R}^{C \times L}$, where $C$ denotes the number of sensor channels and $L$ the sequence length in timesteps. The objective is to predict an activity label $y \in \{1, \dots, K\}$ for $K$ classes. A feature backbone maps the input to a latent sequence $\mathbf{Z} \in \mathbb{R}^{L \times d_{\text{model}}}$, followed by temporal aggregation to produce a fixed dimensional representation for classification.

\subsection{Selective State Space Formulation}

The continuous time linear state space model is defined as:
\begin{align}
\frac{dh(t)}{dt} &= \mathbf{A} h(t) + \mathbf{B} x(t) \\
y(t) &= \mathbf{C} h(t) + \mathbf{D} x(t)
\end{align}
where $h(t) \in \mathbb{R}^N$ is the latent state, and $\mathbf{A} \in \mathbb{R}^{N \times N}$, $\mathbf{B} \in \mathbb{R}^{N \times d}$, $\mathbf{C} \in \mathbb{R}^{d \times N}$, $\mathbf{D} \in \mathbb{R}^{d \times d}$ are learnable parameters. Discretization through zero order hold with step size $\Delta$ yields:
\begin{align}
h_t &= \bar{\mathbf{A}} h_{t-1} + \bar{\mathbf{B}} x_t \\
y_t &= \mathbf{C} h_t + \mathbf{D} x_t
\end{align}
where $\bar{\mathbf{A}} = \exp(\Delta \mathbf{A})$ and $\bar{\mathbf{B}} = (\Delta \mathbf{A})^{-1}(\exp(\Delta \mathbf{A}) - \mathbf{I}) \Delta \mathbf{B}$.

The Mamba selective mechanism \cite{gu2023mamba} makes $\Delta_t$, $\mathbf{B}_t$, and $\mathbf{C}_t$ input dependent:
\begin{align}
\Delta_t &= \text{softplus}(\mathbf{W}_\Delta x_t) \\
\mathbf{B}_t &= \mathbf{W}_B x_t, \quad \mathbf{C}_t = \mathbf{W}_C x_t
\end{align}
This selectivity enables the model to dynamically gate state updates: small $\Delta_t$ preserves history, while large $\Delta_t$ encourages forgetting. For HAR, this allows selective attention to motion transitions while suppressing stationary noise.

\subsection{BabyMamba-HAR Architecture Family}

Two complementary architectures are proposed, addressing distinct sensor configuration scenarios through different stem designs.

\begin{figure*}[t]
\centering
\includegraphics[width=0.75\textwidth]{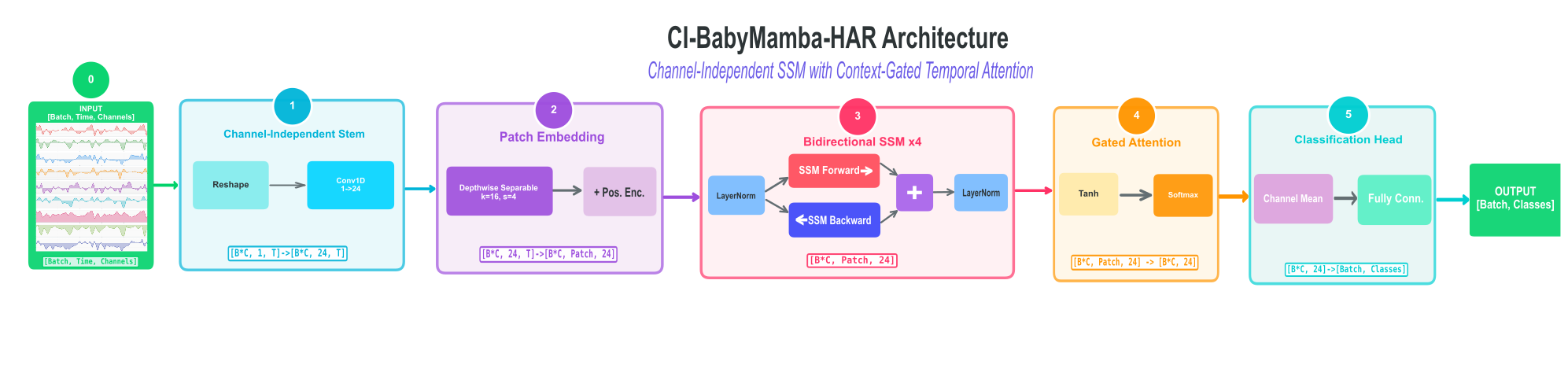}
\caption{CI-BabyMamba-HAR architecture with channel independent stem, weight tied bidirectional SSM blocks, and context gated temporal attention head.}
\label{fig:ci_arch}
\end{figure*}

\subsubsection{CI-BabyMamba-HAR (Channel Independent)}

The channel independent architecture (Fig.~\ref{fig:ci_arch}) addresses noise isolation in heterogeneous sensor arrays. Each sensor channel is processed independently through a shared convolutional stem, enabling noise isolation between heterogeneous sensors while maintaining parameter efficiency through weight sharing. The processed channels flow through stacked weight tied bidirectional SSM blocks, where forward and backward scans share parameters to double the receptive field without increasing model size. This architecture is particularly suited for datasets with heterogeneous or noisy sensor configurations where cross channel interference must be minimized. Each channel $\mathbf{X}^{(c)} \in \mathbb{R}^{1 \times L}$ is processed through a shared convolutional stem:
\begin{equation}
\mathbf{Z}^{(c)} = \sigma(\text{BN}(\text{Conv1d}(\mathbf{X}^{(c)}; \mathbf{W}_{\text{stem}})))
\end{equation}
where $\mathbf{W}_{\text{stem}} \in \mathbb{R}^{d_{\text{model}} \times 1 \times k}$ are shared weights and $\sigma(\cdot)$ is SiLU activation. The backbone processes $B \cdot C$ independent sequences, and late fusion averages channel representations:
\begin{equation}
\mathbf{h} = \frac{1}{C} \sum_{c=1}^{C} \text{pool}(\phi_{\text{ssm}}(\mathbf{Z}^{(c)}))
\end{equation}

\textbf{Configuration:} $d_{\text{model}}=24$, $d_{\text{state}}=16$, $n_{\text{layers}}=4$, expand$=2$, yielding $\sim$28K parameters.

\begin{figure*}[t]
\centering
\includegraphics[width=0.75\textwidth]{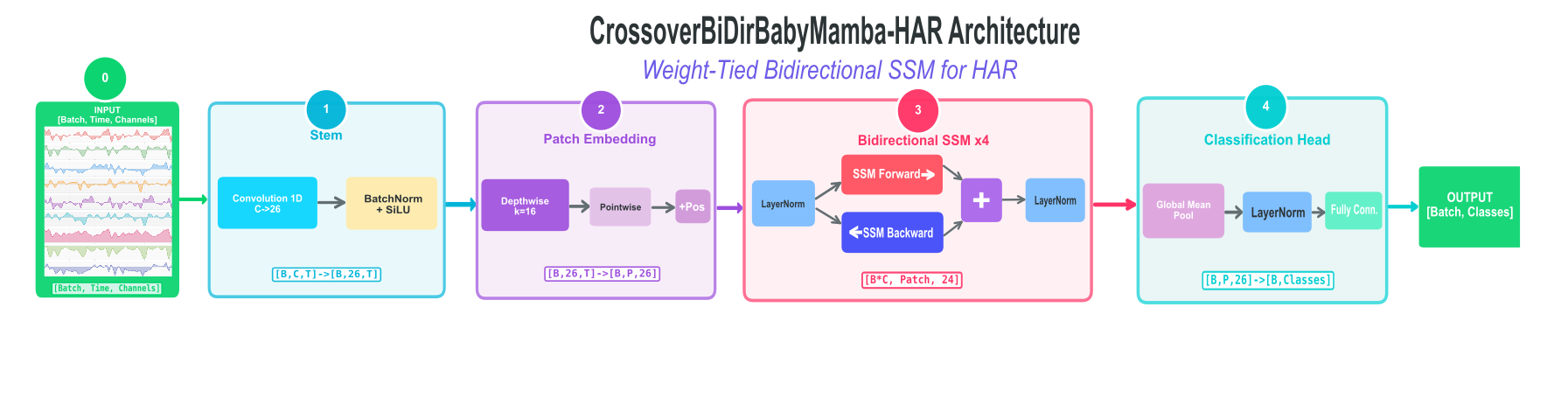}
\caption{Crossover-BiDir-BabyMamba-HAR architecture with early fusion stem and bidirectional SSM blocks with crossover connections.}
\label{fig:crossover_arch}
\end{figure*}

\subsubsection{Crossover-BiDir-BabyMamba-HAR (Early Fusion)}

The early fusion architecture (Fig.~\ref{fig:crossover_arch}) achieves channel count independent complexity through immediate projection. All $C$ input channels are fused into $d_{\text{model}}$ features through a single convolution operation, and the crossover mechanism facilitates information exchange between forward and backward processing paths in the bidirectional SSM blocks. This design is particularly efficient on high channel datasets such as Opportunity (79 channels), where 11$\times$ fewer MACs are achieved compared to conventional architectures while maintaining competitive accuracy.
\begin{equation}
\mathbf{Z} = \sigma(\text{BN}(\text{Conv1d}(\mathbf{X}; \mathbf{W}_{\text{fused}})))
\end{equation}
where $\mathbf{W}_{\text{fused}} \in \mathbb{R}^{d_{\text{model}} \times C \times k}$ maps $C$ channels to $d_{\text{model}}$ dimensions. The backbone processes only $B$ sequences regardless of $C$, yielding:
\begin{equation}
\text{MACs}_{\text{Crossover}} = \frac{\text{MACs}_{\text{CI}}}{C}
\end{equation}

\textbf{Configuration:} $d_{\text{model}}=26$, $d_{\text{state}}=8$, $n_{\text{layers}}=4$, expand$=2$, yielding $\sim$27K parameters.

\begin{figure*}[t]
\centering
\includegraphics[width=0.75\textwidth]{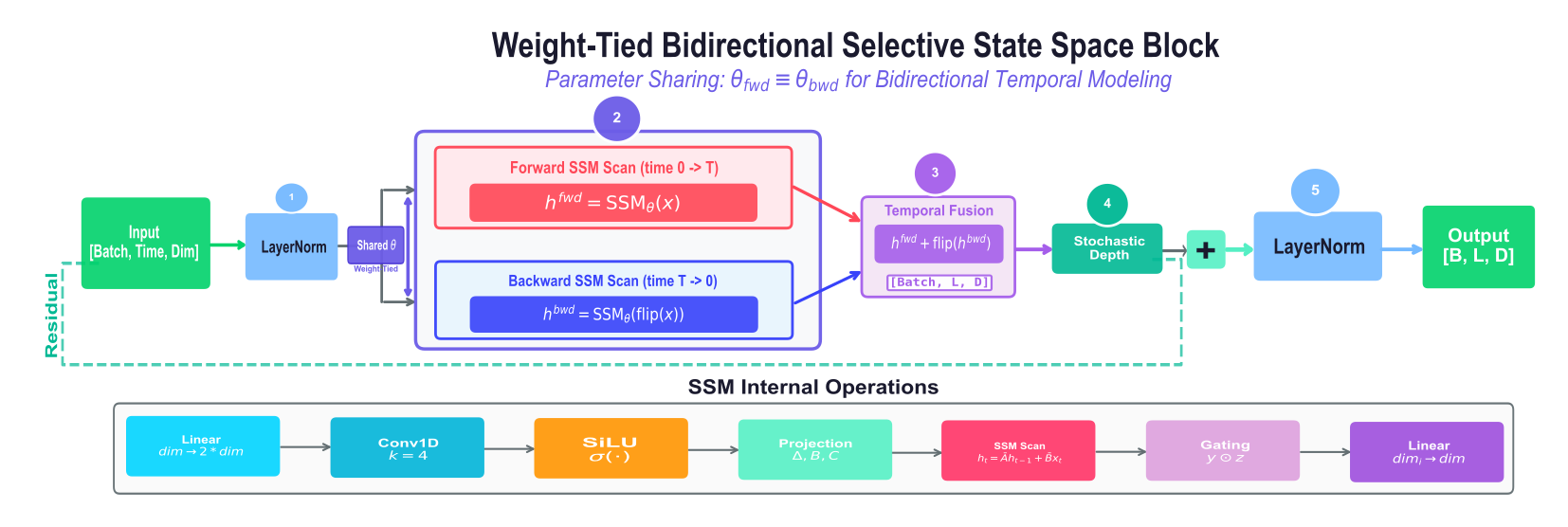}
\caption{Weight tied bidirectional selective state space (SSM) block architecture with input dependent projections and parallel scan implementation.}
\label{fig:ssm_block}
\end{figure*}

\subsubsection{Weight Tied Bidirectional Scanning}

Windowed HAR classification has access to the complete input sequence, eliminating causality constraints. Bidirectional context is introduced through weight tied forward and backward scans (Fig.~\ref{fig:ssm_block}). The Mamba inspired selective mechanism makes the discretization step $\Delta_t$ and state matrices $\mathbf{B}_t$, $\mathbf{C}_t$ input dependent through learned linear projections, enabling dynamic control of information flow: small $\Delta_t$ values preserve historical context while large values encourage selective forgetting. This content aware gating is particularly beneficial for HAR, where discriminative motion transitions are brief while stationary periods contribute primarily noise. The formulation is as follows:
\begin{align}
\mathbf{H}_{\text{fwd}} &= \text{SSM}_\theta(\mathbf{Z}) \\
\mathbf{H}_{\text{bwd}} &= \mathcal{T}(\text{SSM}_\theta(\mathcal{T}(\mathbf{Z}))) \\
\mathbf{Z}_{\text{out}} &= \text{LN}(\mathbf{Z} + \mathbf{H}_{\text{fwd}} + \mathbf{H}_{\text{bwd}})
\end{align}
where $\mathcal{T}$ denotes temporal reversal and $\theta$ are shared parameters. This doubles the effective receptive field without parameter increase.

\subsubsection{Context Gated Temporal Attention Pooling}

Global average pooling treats all timesteps equally, diluting brief discriminative segments. Learnable attention is employed:
\begin{align}
\mathbf{e}_t &= \tanh(\mathbf{W}_g \mathbf{z}_t + \mathbf{b}_g) \\
\alpha_t &= \frac{\exp(\mathbf{v}^\top \mathbf{e}_t)}{\sum_{j=1}^{L} \exp(\mathbf{v}^\top \mathbf{e}_j)} \\
\mathbf{c} &= \sum_{t=1}^{L} \alpha_t \mathbf{z}_t
\end{align}
This adds only $d_{\text{model}} \times d_{\text{attn}} + 2 d_{\text{attn}} \approx 624$ parameters.

\subsection{Computational Complexity}

The SSM backbone has complexity $O(L \cdot d_{\text{model}} \cdot d_{\text{state}})$, linear in sequence length. For Crossover-BiDir-BabyMamba-HAR, backbone compute is independent of $C$. For CI-BabyMamba-HAR, complexity scales as $O(L \cdot C \cdot d_{\text{model}} \cdot d_{\text{state}})$. This distinction is critical for high channel datasets.

\section{Experimental Setup}

\subsection{Evaluation Protocol}

To ensure fair comparison, all models share identical data loading pipelines, metric computation (macro F1-score), and random seed control. Subject independent splits followed dataset author specifications where available, otherwise, leave one subject out cross-validation (LOSO CV) was applied. Results are reported as mean $\pm$ standard deviation over five seeds generated from a master seed.

A critical methodological consideration arises for single subject datasets where standard subject-wise splitting is inapplicable. The Skoda dataset~\cite{skoda} presents a particularly instructive case: with only one participant performing industrial assembly activities in a manufacturing environment, naive random splitting of overlapping windows induces severe data leakage, as adjacent windows share 75\% of their samples. To address this, a \emph{temporal split} is used strategy within each activity class, the first 80\% of chronologically ordered windows are assigned to training and the remaining 20\% to testing. This protocol ensures that test samples are temporally disjoint from training data, preventing artificially inflated performance metrics that would otherwise arise from near duplicate samples spanning the train-test boundary.

\subsection{Dataset Specific Preprocessing}

Dataset preprocessing was tailored to sensor characteristics:

\begin{itemize}
\item \textbf{Standard preprocessing} (UCI-HAR, MotionSense, WISDM, Opportunity, UniMiB, Daphnet): Per-channel z-score normalization ($x' = (x - \mu_c) / \sigma_c$) computed on training data.
\item \textbf{Signal rescue} (PAMAP2): Robust scaling (median/IQR) to handle hardware artifacts from intentionally loose sensors, combined with 5~Hz Butterworth low-pass filtering to remove high frequency noise while preserving activity signatures.
\item \textbf{Signal rescue} (Skoda): 5~Hz low-pass Butterworth filter to suppress industrial vibration artifacts from the assembly line environment, followed by z-score normalization. Overlapping windows (75\% overlap, stride=24 for 98-sample windows) are employed to augment the limited single-subject data.
\end{itemize}

All filtering used 4th order Butterworth design with zero phase forward-backward application to prevent temporal distortion.

\subsection{Datasets}

Eight public benchmarks spanning diverse sensor modalities were evaluated (Table~\ref{tab:datasets}).

\begin{table}[t]
\caption{Dataset characteristics and windowing configuration.}
\label{tab:datasets}
\centering
\scriptsize
\setlength{\tabcolsep}{3pt}
\begin{tabular}{lcccccc}
\toprule
Dataset & Subj. & Classes & Ch. & Hz & $L$ & Sec. \\
\midrule
UCI-HAR \cite{uci_har} & 30 & 6 & 9 & 50 & 128 & 2.56 \\
MotionSense \cite{motionsense} & 24 & 6 & 6 & 50 & 128 & 2.56 \\
WISDM \cite{wisdm} & 36 & 6 & 3 & 20 & 128 & 6.40 \\
PAMAP2 \cite{pamap2} & 9 & 12 & 19 & 100 & 128 & 1.28 \\
Opportunity \cite{opportunity} & 4 & 5 & 79 & 30 & 128 & 4.27 \\
UniMiB-SHAR \cite{unimib} & 30 & 9 & 3 & 50 & 128 & 2.56 \\
Skoda \cite{skoda} & 1 & 11 & 30 & 98 & 98 & 1.00 \\
Daphnet \cite{daphnet} & 10 & 2 & 9 & 64 & 64 & 1.00 \\
\bottomrule
\end{tabular}
\end{table}

\subsection{Training Configuration}

Models were trained with AdamW optimizer \cite{loshchilov2019adamw} ($\beta_1=0.9$, $\beta_2=0.999$), learning rate scheduler (ReduceLROnPlateau, factor=0.5, patience=5), and early stopping (patience=10 on validation F1-score). Maximum epochs: 200. Gradient clipping: max\_norm=1.0. Loss: CrossEntropy with label smoothing $\epsilon=0.1$.

Online data augmentation included time warping ($p=0.5$), magnitude scaling ($p=0.5$, $\alpha \sim \mathcal{U}(0.8, 1.2)$), Gaussian jitter ($p=0.3$, $\sigma=0.05$), and channel dropout ($p=0.2$).

\subsection{Baselines}

All baselines were re-implemented in the unified codebase: TinyHAR \cite{zhou2022tinyhar} ($\sim$55K params), TinierHAR \cite{tinierhar2025} ($\sim$33K params), and DeepConvLSTM \cite{ordonez2016deep} ($\sim$136K params).

\begin{figure*}[t]
\centering
\includegraphics[height=0.28\textheight,width=0.85\textwidth,keepaspectratio]{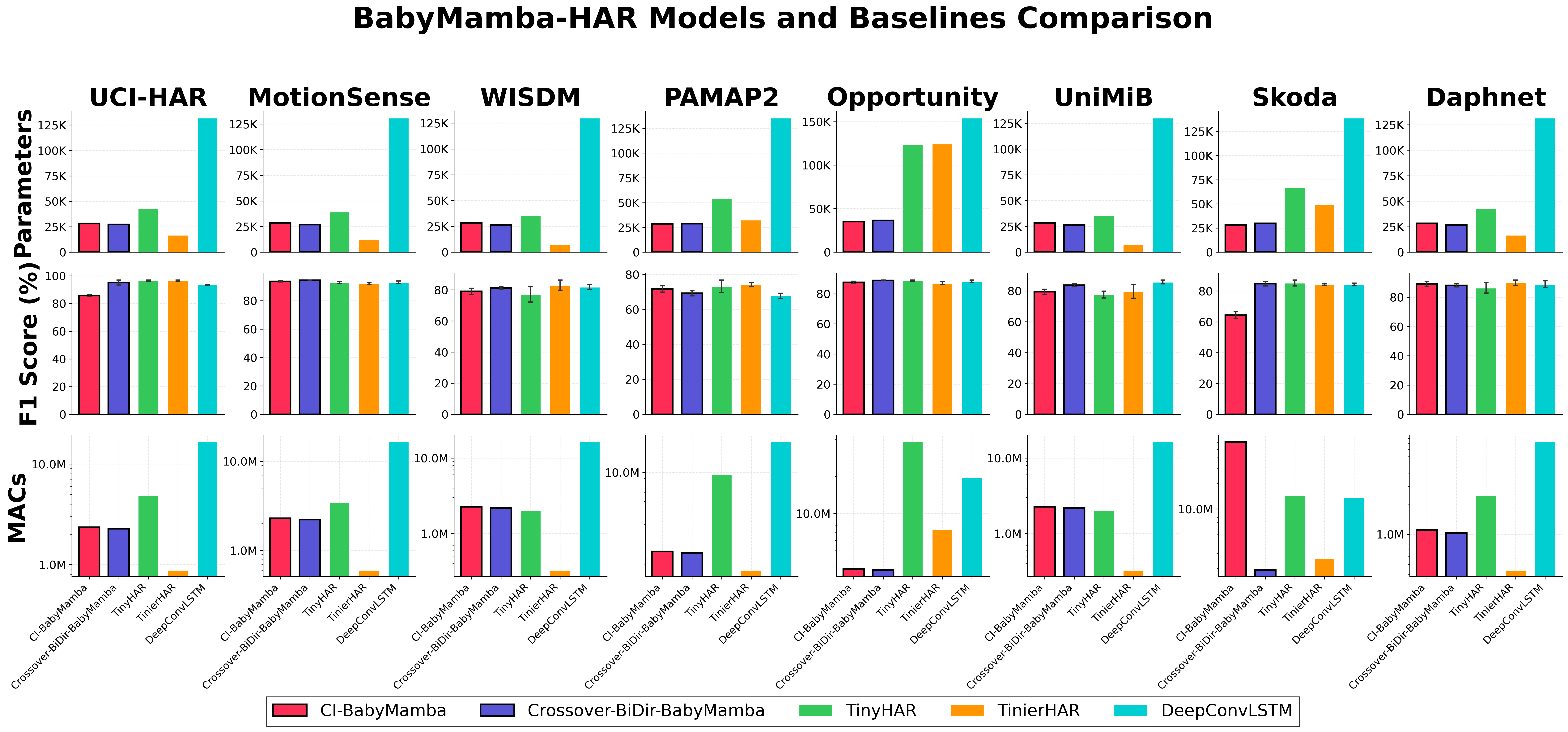}
\caption{Performance comparison grid showing macro F1-scores across all eight HAR benchmark datasets.}
\label{fig:results_grid}
\end{figure*}

\section{Results}

\subsection{Overall Performance Comparison}

Table~\ref{tab:main_results} presents macro F1-scores across all datasets. Crossover-BiDir-BabyMamba-HAR achieves 86.52\% average F1-score, comparable to TinyHAR (86.16\%), and within 0.87 points of TinierHAR (87.39\%), while maintaining $\sim$27K parameters.

\begin{table*}[t]
\caption{Macro F1-score (\%) comparison across all datasets. Best result per dataset in \textbf{bold}.}
\label{tab:main_results}
\centering
\scriptsize
\setlength{\tabcolsep}{4pt}
\begin{tabular}{l|cccccccc|c}
\toprule
Model & UCI-HAR$\uparrow$ & Motion.$\uparrow$ & WISDM$\uparrow$ & PAMAP2$\uparrow$ & Opport.$\uparrow$ & UniMiB$\uparrow$ & Skoda$\uparrow$ & Daphnet$\uparrow$ & \textbf{Avg}$\uparrow$ \\
\midrule
Crossover-BiDir-BabyMamba & 95.13{\tiny$\pm$1.79} & \textbf{93.03}{\tiny$\pm$0.49} & 80.69{\tiny$\pm$1.44} & 65.67{\tiny$\pm$2.71} & \textbf{88.81}{\tiny$\pm$0.28} & 83.74{\tiny$\pm$0.93} & 84.74{\tiny$\pm$1.50} & 88.08{\tiny$\pm$1.07} & 85.02 \\
CI-BabyMamba-HAR & 84.79{\tiny$\pm$0.39} & 86.21{\tiny$\pm$1.50} & 77.76{\tiny$\pm$2.40} & 65.46{\tiny$\pm$2.52} & 76.73{\tiny$\pm$1.76} & 80.83{\tiny$\pm$2.97} & 64.34{\tiny$\pm$2.23} & 84.59{\tiny$\pm$0.79} & 77.59 \\
TinyHAR \cite{zhou2022tinyhar} & \textbf{96.53}{\tiny$\pm$0.41} & 92.67{\tiny$\pm$0.67} & 77.09{\tiny$\pm$4.95} & 73.22{\tiny$\pm$3.58} & 88.69{\tiny$\pm$0.38} & 77.61{\tiny$\pm$2.23} & \textbf{85.22}{\tiny$\pm$1.91} & 86.42{\tiny$\pm$3.64} & 86.01 \\
TinierHAR \cite{tinierhar2025} & 96.37{\tiny$\pm$0.57} & 91.99{\tiny$\pm$0.60} & \textbf{83.06}{\tiny$\pm$3.24} & \textbf{74.07}{\tiny$\pm$1.16} & 87.09{\tiny$\pm$0.90} & 79.67{\tiny$\pm$4.45} & 84.21{\tiny$\pm$0.47} & \textbf{89.84}{\tiny$\pm$1.90} & \textbf{85.85} \\
DeepConvLSTM \cite{ordonez2016deep} & 93.53{\tiny$\pm$0.26} & 92.90{\tiny$\pm$0.96} & 81.84{\tiny$\pm$1.46} & 67.79{\tiny$\pm$1.50} & 88.30{\tiny$\pm$0.72} & \textbf{85.83}{\tiny$\pm$1.22} & 84.27{\tiny$\pm$0.88} & 88.95{\tiny$\pm$2.26} & 85.42 \\
\bottomrule
\end{tabular}
\end{table*}

\begin{table}[t]
\caption{Computational efficiency comparison (averaged across datasets).}
\label{tab:efficiency}
\centering
\scriptsize
\setlength{\tabcolsep}{4pt}
\begin{tabular}{lccc}
\toprule
Model & Params$\downarrow$ & Avg MACs$\downarrow$ & F1/M-MACs$\uparrow$\\
\midrule
Crossover-BiDir-BabyMamba & \textbf{27K} & 2.21M & 39.1 \\
CI-BabyMamba-HAR & 28K & 50.92M & 1.6 \\
TinyHAR & 55K & 9.29M & 9.3 \\
TinierHAR & 33K & \textbf{1.73M} & \textbf{50.5} \\
DeepConvLSTM & 136K & 15.51M & 5.5 \\
\bottomrule
\end{tabular}
\end{table}

\subsection{Computational Efficiency}

Table~\ref{tab:efficiency} summarizes computational costs. Crossover-BiDir-BabyMamba-HAR achieves 11$\times$ fewer MACs than TinyHAR on Opportunity (3.44M vs. 38.30M MACs) while matching F1-score (88.81\% vs. 88.69\%). CI-BabyMamba-HAR's MACs scale with channel count, reaching 222M MACs on Opportunity (79 channels), making it impractical for high dimensional datasets.

\subsection{High Channel Dataset Analysis}

On Opportunity (79 channels) and Skoda (30 channels), the computational advantage of the Crossover-BiDir architecture is quantified in Table~\ref{tab:high_channel}:

\begin{table}[t]
\caption{Performance on high channel datasets.}
\label{tab:high_channel}
\centering
\scriptsize
\setlength{\tabcolsep}{3pt}
\begin{tabular}{llccc}
\toprule
Dataset & Model & F1 (\%)$\uparrow$ & MACs$\downarrow$ & Speedup$\uparrow$ \\
\midrule
\multirow{3}{*}{Opportunity} & Crossover-BabyMamba & \textbf{88.81} & \textbf{3.44M} & \textbf{11.1$\times$} \\
& TinyHAR & 88.69 & 38.30M & 1.0$\times$ \\
& CI-BabyMamba & 76.73 & 222.31M & 0.17$\times$ \\
\midrule
\multirow{3}{*}{Skoda} & Crossover-BabyMamba & 84.74 & \textbf{1.92M} & \textbf{6.0$\times$} \\
& TinyHAR & \textbf{85.22} & 11.48M & 1.0$\times$ \\
& DeepConvLSTM & 84.27 & 13.39M & 0.86$\times$ \\
\bottomrule
\end{tabular}
\end{table}

A visualization of performance across all eight datasets is presented in Fig.~\ref{fig:results_grid}. As shown, Crossover-BiDir-BabyMamba-HAR demonstrates consistent strong performance across diverse sensor configurations, achieving the highest F1-scores on MotionSense (93.03\%) and Opportunity (88.81\%). Notably, on Skoda, where the temporal splitting protocol eliminates data leakage from overlapping windows, TinyHAR achieves 85.22\% F1-score compared to Crossover-BiDir-BabyMamba's 84.74\%. This modest gap contrasts sharply with pre-correction results where both models exceeded 97\%, underscoring the importance of rigorous evaluation protocols for single-subject datasets. The contrasting performance profiles between CI-BabyMamba-HAR and Crossover-BiDir-BabyMamba-HAR show the importance of stem architecture selection based on dataset characteristics: early fusion excels when channels exhibit high correlation, while channel independent processing provides robustness for heterogeneous sensor arrays.

\begin{figure*}[t]
\centering
\includegraphics[height=0.28\textheight,width=0.85\textwidth,keepaspectratio]{babymamba_ablation_combined.png}
\caption{Combined ablation study results showing $\Delta$F1 relative to baseline configurations. (C)~Channel processing, (B)~hyperparameter sensitivity, (A)~architecture ablation, and (L)~sequence length scaling. Yellow/gold bars: CI-BabyMamba-HAR, purple bars: Crossover-BabyMamba-HAR.}
\label{fig:ablation_combined}
\end{figure*}

\section{Ablation Studies}

Systematic ablations isolate architectural contributions using controlled single variable experiments. All ablations report mean $\pm$ std over 5 seeds.

\subsection{Bidirectionality Impact}

\begin{table}[t]
\caption{Ablation: Bidirectional vs. unidirectional scanning.}
\label{tab:ablation_bidir}
\centering
\scriptsize
\setlength{\tabcolsep}{3pt}
\begin{tabular}{llcc}
\toprule
Dataset & Model & BiDir F1 (\%) & UniDir F1 (\%) \\
\midrule
\multirow{2}{*}{MotionSense} & CI-BabyMamba-HAR& 93.47 & 85.05 (\textbf{-8.42}) \\
& Crossover-BabyMamba& 94.31 & 92.74 (-1.57) \\
\midrule
\multirow{2}{*}{PAMAP2} & CI-BabyMamba-HAR& 71.67 & 65.35 (\textbf{-6.32}) \\
& Crossover-BabyMamba & 69.20 & 64.36 (-4.84) \\
\midrule
\multirow{2}{*}{UCI-HAR} & CI-BabyMamba-HAR& 85.80 & 85.88 (+0.08) \\
& Crossover-BabyMamba & 95.11 & 95.10 (-0.01) \\
\bottomrule
\end{tabular}
\end{table}

Table~\ref{tab:ablation_bidir} demonstrates that bidirectional scanning is critical for complex temporal patterns (MotionSense: -8.42\% F1-score, PAMAP2: -6.32\% F1-score), while simpler datasets (UCI-HAR) show minimal impact. The Crossover architecture is more robust to unidirectional ablation, likely due to the crossover mechanism providing implicit redundancy.

\subsection{Temporal Pooling Strategy}

\begin{table}[t]
\caption{Ablation: Gated attention vs. mean pooling.}
\label{tab:ablation_pooling}
\centering
\scriptsize
\setlength{\tabcolsep}{3pt}
\begin{tabular}{llcc}
\toprule
Dataset & Model & Gated Attn F1 (\%) & Mean Pool F1 (\%) \\
\midrule
\multirow{2}{*}{MotionSense} & CI-BabyMamba & 93.47 & 84.53 (\textbf{-8.94}) \\
& Crossover-BabyMamba& 94.31 & 93.11 (-1.20) \\
\midrule
\multirow{2}{*}{PAMAP2} & CI-BabyMamba & 71.67 & 67.98 (-3.69) \\
& Crossover-BabyMamba & 69.20 & 66.39 (-2.81) \\
\bottomrule
\end{tabular}
\end{table}

Gated temporal attention provides considerable improvements for CI-BabyMamba-HAR (Table~\ref{tab:ablation_pooling}), with up to 8.94\% F1-score gain on MotionSense. This confirms that learnable attention is essential for focusing on discriminative temporal segments.

\subsection{Channel Processing Strategy}

The C2 ablation evaluating explicit channel independent processing was completed across all eight datasets for both architectures (80 total runs: 8 datasets $\times$ 2 models $\times$ 5 seeds). Table~\ref{tab:ablation_channel} summarizes key findings.

\begin{table}[t]
\caption{Ablation: Channel processing across all datasets (C2 variant).}
\label{tab:ablation_channel}
\centering
\scriptsize
\setlength{\tabcolsep}{2.5pt}
\begin{tabular}{lccl}
\toprule
Dataset & CI-BabyMamba & Crossover-BiDir& Winner \\
\midrule
UCI-HAR & 85.17\% & \textbf{94.66\%} & Crossover-BiDir (+9.49)\\
MotionSense & 86.06\% & \textbf{93.05\%} & Crossover-BiDir (+6.99)\\
WISDM & 77.85\% & \textbf{80.78\%} & Crossover-BiDir (+2.93)\\
PAMAP2 & \textbf{65.64\%} & 65.50\% & CI-BabyMamba (+0.14) \\
Opportunity & 77.86\% & \textbf{88.30\%} & Crossover-BiDir (+10.44)\\
UniMiB & 80.28\% & \textbf{83.89\%} & Crossover-BiDir (+3.61)\\
Skoda & 65.60\% & \textbf{84.74\%} & Crossover-BiDir (+19.14)\\
Daphnet & 84.60\% & \textbf{88.00\%} & Crossover-BiDir (+3.40)\\
\bottomrule
\end{tabular}
\end{table}

Crossover's early fusion dominates on 7 out of 8 datasets, with gains of +19\% on Skoda and +10\% on Opportunity. The sole exception is PAMAP2, where intentional sensor artifacts favor channel isolation. This evaluation confirms that stem architecture should prioritize fusion for correlated sensors and isolation only when independent noise profiles exist.

\subsection{Hyperparameter Sensitivity}

\begin{table}[t]
\caption{Ablation: Hyperparameter sensitivity summary.}
\label{tab:ablation_hyper}
\centering
\scriptsize
\setlength{\tabcolsep}{3pt}
\begin{tabular}{lcc}
\toprule
Parameter & Avg $\Delta$F1 & $\Delta$Params \\
\midrule
$d_{\text{state}}$: 16 $\rightarrow$ 8 & $<$0.5\% & -16\% \\
$d_{\text{model}}$: 26 $\rightarrow$ 24 & $<$0.5\% & -12\% \\
expand: 2 $\rightarrow$ 3 & $<$1\% & +44\% \\
seq\_len: 128 $\rightarrow$ 64 & -2\% & 0\% \\
\bottomrule
\end{tabular}
\end{table}

Table~\ref{tab:ablation_hyper} indicates remarkable robustness to hyperparameter variation. Reducing $d_{\text{state}}$ from 16 to 8 saves 16\% parameters with $<$0.5\% F1-score impact, enabling sub-25K parameter deployments. Increasing expansion factor to 3 adds 44\% parameters without benefit, confirming that lightweight models are not capacity limited.

\subsection{Key Ablation Findings}

The ablation results are visualized in Fig.~\ref{fig:ablation_combined}. Key findings:

\begin{enumerate}
\item \textbf{Bidirectionality is essential}: Up to 8.42\% F1-score gain on MotionSense. Activities with asymmetric temporal signatures (sit to stand versus stand to sit) benefit from both forward and backward context.
\item \textbf{Gated attention is critical for CI-BabyMamba-HAR}: Up to 8.94\% F1-score improvement. Channel independent processing requires learnable aggregation to focus on discriminative timesteps.
\item \textbf{Early fusion is generally superior}: C2 ablation across all 8 datasets shows Crossover-BiDir-BabyMamba-HAR outperforming CI-BabyMamba-HAR on 7/8 benchmarks (Table~\ref{tab:ablation_channel}), often by large margins (+19\% Skoda, +10\% Opportunity). Only PAMAP2 shows marginal CI advantage (+0.14\%), suggesting channel isolation benefits only datasets with explicit sensor artifacts.
\item \textbf{Hyperparameters are robust}: $d_{\text{state}}$ reduction (16$\rightarrow$8) saves 16\% parameters at $<$0.5\% F1-score cost. Expand factor increase provides no benefit despite +44\% parameters.
\end{enumerate}

\section{Edge Deployment}

To validate that the BabyMamba-HAR architectures remain viable beyond desktop profiling, on-device deployment was done on two representative microcontroller platforms. The deployment path was implemented through a handcrafted recurrent C++ runtime rather than through a graph compiler such as TFLite Micro. This design choice was motivated by the observation that the selective state space recurrence can be expressed as a tight sequential loop that avoids the overhead of a general purpose inference engine.

\subsection{Deployment Pipeline}

The deployment path was organized as a direct checkpoint to header export followed by native compilation and on-device benchmarking. Trained PyTorch checkpoints were serialized into C++ header arrays through a custom export script. Projection heavy matrices were compressed into row wise INT8 storage with per row float32 scales, while the hidden state evolution, normalization, and recurrent scan path were retained in float32 to preserve numerical parity. This mixed precision strategy targets the projection matrices that dominate flash traffic and matrix-vector cost, while protecting the numerically sensitive recurrent accumulation path.

\subsection{Operator Fusion and Memory Management}

The BabyMamba-HAR deployment runtime adopts a fused streaming selective scan implementation in the spirit of MambaLite-Micro, but extends it to two HAR specific settings: a weight-tied bidirectional BabyMamba block, and a channel streaming channel independent BabyMamba block. In the native ESP32 runtime, this fused recurrent execution is further combined with projection only INT8 storage and dual core CI execution.

The reference PyTorch implementation relies on materialization of intermediate tensors for the discretization and recurrent update steps. Specifically, the discretization step constructs the multidimensional tensors for all time steps before running the chunked scan, which demands a high memory footprint. To address this limitation, the BabyMamba-HAR runtime uses a fused computation strategy. By fusing the discretization and the sequential update into a single streaming operation, the state evolution is computed iteratively on the fly inside the token loop. This operator fusion eliminates the need to materialize the full sequence tensors. Table~\ref{tab:algorithm_comparison} presents a comparison between the reference PyTorch style sequence processing and the current deployed fused implementation.

\begin{table*}[t]
\caption{Algorithmic comparison between original PyTorch style sequence modeling and the proposed BabyMamba fused embedded implementation.}
\label{tab:algorithm_comparison}
\centering
\scriptsize
\begin{tabular}{p{0.45\textwidth} | p{0.45\textwidth}}
\toprule
\textbf{Algorithm 1:} Original BabyMamba PyTorch Style Selective Scan& \textbf{Algorithm 2:} Current BabyMamba Fused Embedded Selective Scan \\
\midrule
\textbf{Input:} Token sequence $\mathbf{X}$ \newline
\textbf{Output:} Sequence $\mathbf{Y}$ \newline
\newline
\textit{Explicit Materialization (High Memory Requirement):} \newline
1: $\mathbf{U} \leftarrow \text{InProj}(\mathbf{X})$ \newline
2: Split $\mathbf{U}$ into $\mathbf{X}_{\text{part}}$ and $\mathbf{Z}_{\text{part}}$ \newline
3: $\mathbf{V} \leftarrow \text{Conv1d}(\mathbf{X}_{\text{part}})$ \newline
4: $\mathbf{A}_{\text{act}} \leftarrow \text{SiLU}(\mathbf{V})$ \newline
5: $\mathbf{P} \leftarrow \text{XProj}(\mathbf{A}_{\text{act}})$ \newline
6: Split $\mathbf{P}$ into $\mathbf{\Delta}_{\text{base}}$, $\mathbf{B}$, $\mathbf{C}$ \newline
7: $\mathbf{\Delta} \leftarrow \text{Softplus}(\text{DtProj}(\mathbf{\Delta}_{\text{base}}))$ \newline
8: $\mathbf{\Delta}_{\mathbf{A}} \leftarrow \exp(\mathbf{\Delta} \otimes \mathbf{A}_{\text{log}})$ \newline
9: $\mathbf{\Delta}_{\mathbf{BX}} \leftarrow \mathbf{\Delta} \otimes \mathbf{B} \otimes \mathbf{A}_{\text{act}}$ \newline
\newline
\textit{Sequential Scan:} \newline
10: $\mathbf{H} \leftarrow \text{Scan}(\mathbf{\Delta}_{\mathbf{A}}, \mathbf{\Delta}_{\mathbf{BX}})$ \newline
11: $\mathbf{Y}_{\text{inner}} \leftarrow \mathbf{C} \otimes \mathbf{H} + \mathbf{D} \otimes \mathbf{A}_{\text{act}}$ \newline
12: $\mathbf{Y} \leftarrow \text{OutProj}(\mathbf{Y}_{\text{inner}} \odot \text{SiLU}(\mathbf{Z}_{\text{part}}))$ \newline
13: \textbf{return} $\mathbf{Y}$
&
\textbf{Input:} Token sequence $\mathbf{X}$ \newline
\textbf{Output:} Sequence $\mathbf{Y}$ \newline
\newline
\textit{Fused Streaming Execution (Low Memory Requirement):} \newline
1: Initialize hidden state and local convolution history buffers \newline
2: \textbf{for} each token $x_t$ \textbf{do} \newline
3: \quad Compute $x_{\text{part}, t}$ and $z_{\text{part}, t}$ directly from $\text{InProj}(x_t)$ \newline
4: \quad Update convolution history in place \newline
5: \quad Compute activated conv response $a_t$ \newline
6: \quad Compute $x_{\text{proj\_full}, t}$ from $a_t$ \newline
7: \quad Compute $dt_t$ from the $dt$ subvector of $x_{\text{proj\_full}, t}$ \newline
8: \quad \textbf{for} each inner channel $i$ and state element $n$ \textbf{do} \newline
9: \quad \quad Read $B_t(i, n)$ and $C_t(i, n)$ from $x_{\text{proj\_full}, t}$ \newline
10: \quad \quad Update $h_t(i, n)$ directly \newline
11: \quad \quad Accumulate $y_{\text{inner}, t}(i)$ directly \newline
12: \quad \textbf{end for} \newline
13: \quad Apply the $z$ gate and OutProj immediately \newline
14: \textbf{end for} \newline
15: \textbf{return} $\mathbf{Y}$\\
\bottomrule
\end{tabular}
\end{table*}

Memory usage is optimized through lifetime aware buffer allocation, specifically customized for the variant level execution. For Crossover-BiDir-BabyMamba-HAR, the fused scan operates inside a weight-tied bidirectional residual block under a fixed scratch-buffer regime (Table~\ref{tab:crossover_algorithm}). For CI-BabyMamba-HAR, the runtime streams one sensor channel at a time, bypassing the memory intensive, batch expanded activation pipeline altogether (Table~\ref{tab:ci_algorithm}). This strategy eliminates intermediate storage bottlenecks, and enables seamless execution across resource constrained embedded targets.

\begin{table*}[t]
\caption{Algorithmic comparison for Crossover-BiDir-BabyMamba-HAR: PyTorch reference versus deployed runtime.}
\label{tab:crossover_algorithm}
\centering
\scriptsize
\begin{tabular}{p{0.45\textwidth} | p{0.45\textwidth}}
\toprule
\textbf{Algorithm 3:} Original PyTorch Crossover-BiDir Forward & \textbf{Algorithm 4:} Deployed Crossover-BiDir Runtime \\
\midrule
\textbf{Input:} Multichannel window $\mathbf{X}$ \newline
\textbf{Output:} Logits \newline
\newline
1: Compute stem features over all channels jointly \newline
2: Compute patch depthwise convolution \newline
3: Compute patch pointwise projection \newline
4: Normalize and apply SiLU \newline
5: Add positional embedding \newline
6: \textbf{for} each BabyMamba layer \textbf{do} \newline
7: \quad Pre-normalize sequence \newline
8: \quad Run forward selective scan with SSM parameters \newline
9: \quad Flip sequence \newline
10: \quad Run backward selective scan with the same SSM parameters \newline
11: \quad Flip the backward result back \newline
12: \quad Add forward and backward outputs \newline
13: \quad Add residual and post-normalize \newline
14: \textbf{end for} \newline
15: Mean pool the final sequence \newline
16: Normalize and classify&
\textbf{Input:} Multichannel window $\mathbf{X}$ \newline
\textbf{Output:} Logits \newline
\newline
1: Compute stem features directly into scratch \newline
2: Compute patch embedding into scratch \newline
3: \textbf{for} each BabyMamba layer \textbf{do} \newline
4: \quad Layer normalize current scratch sequence \newline
5: \quad Run fused forward selective scan into forward scratch \newline
6: \quad Run fused reverse selective scan into backward scratch \newline
7: \quad Combine residual, forward, and backward responses in scratch \newline
8: \quad Post normalize in scratch \newline
9: \textbf{end for} \newline
10: Mean pool from scratch \newline
11: Normalize and classify\\
\bottomrule
\end{tabular}
\end{table*}

\begin{table*}[t]
\caption{Algorithmic comparison for CI-BabyMamba-HAR: PyTorch reference versus deployed runtime.}
\label{tab:ci_algorithm}
\centering
\scriptsize
\begin{tabular}{p{0.45\textwidth} | p{0.45\textwidth}}
\toprule
\textbf{Algorithm 5:} Original PyTorch CI-BabyMamba Forward & \textbf{Algorithm 6:} Deployed CI-BabyMamba Runtime \\
\midrule
\textbf{Input:} Multichannel window $\mathbf{X}$ with shape $[B, L, C]$ \newline
\textbf{Output:} Logits \newline
\newline
1: Reshape $\mathbf{X}$ into a channel independent batch $[B \cdot C, 1, L]$ \newline
2: Run the shared CI stem for all channels \newline
3: Run patch depthwise convolution for all channel samples \newline
4: Run patch pointwise projection for all channel samples \newline
5: Normalize, activate, and transpose \newline
6: Add positional embedding \newline
7: \textbf{for} each BabyMamba layer \textbf{do} \newline
8: \quad Run the weight-tied bidirectional BabyMamba block \newline
9: \textbf{end for} \newline
10: Run gated temporal attention on all channel samples \newline
11: Reshape back to $[B, C, D]$ \newline
12: Mean pool across channels \newline
13: Normalize and classify&
\textbf{Input:} Multichannel window $\mathbf{X}$ \newline
\textbf{Output:} Logits \newline
\newline
1: Initialize a channel accumulation vector \newline
2: \textbf{for} each sensor channel $c$ \textbf{do} \newline
3: \quad Run the single channel stem into scratch \newline
4: \quad Run patch embedding into scratch \newline
5: \quad \textbf{for} each BabyMamba layer \textbf{do} \newline
6: \quad \quad Run fused forward selective scan \newline
7: \quad \quad Run fused reverse selective scan \newline
8: \quad \quad Combine residual and normalize in scratch \newline
9: \quad \textbf{end for} \newline
10: \quad Run gated temporal attention for this channel \newline
11: \quad Accumulate the pooled channel descriptor \newline
12: \textbf{end for} \newline
13: Divide the accumulated descriptor by channel count \newline
14: Normalize and classify\\
\bottomrule
\end{tabular}
\end{table*}

\subsection{Target Platforms}

Two platforms were selected to span the commonly available microcontroller design space for wearable HAR.

\textbf{Raspberry Pi Pico~2} (RP2350): Dual core Arm Cortex-M33 at 150~MHz, 520~KB SRAM, 4~MB flash. The Pico~2 runtime was compiled with the Pico SDK and executed with a fixed serial benchmark harness. Ten timed inference iterations were recorded after a warm-up pass for each dataset bundle.

\textbf{ESP32} (ESP32-D0WD-V3): Dual core Xtensa LX6 at 240~MHz, 520~KB SRAM, 4~MB flash, no PSRAM. The ESP32 runtime was compiled with the native ESP-IDF toolchain at \texttt{-O3} optimization. Dual core channel splitting was applied to CI-BabyMamba-HAR to distribute the per-channel recurrence across both cores.

\subsection{Deployment Results}

Both BabyMamba families achieved full 8/8 dataset coverage on both platforms. Table~\ref{tab:deploy_family} presents the family level summary.

\begin{table}[t]
\caption{Cross platform deployment summary for BabyMamba-HAR families.}
\label{tab:deploy_family}
\centering
\scriptsize
\setlength{\tabcolsep}{2.5pt}
\begin{tabular}{llcccc}
\toprule
Platform & Family & Runs$\uparrow$ & Latency (ms)$\downarrow$ & Parity (\%)$\uparrow$ & Flash (KB)$\downarrow$ \\
\midrule
\multirow{2}{*}{Pico~2} & Crossover-BiDir & \textbf{8/8} & \textbf{481.9} & \textbf{99.98} & \textbf{171--247} \\
 & CI-BabyMamba & \textbf{8/8} & 11762.0 & 99.94 & 178--217 \\
\midrule
\multirow{2}{*}{ESP32} & Crossover-BiDir & \textbf{8/8} & \textbf{154.4} & 99.20 & \textbf{203--279} \\
 & CI-BabyMamba & \textbf{8/8} & 2768.1 & \textbf{99.36} & 216--254 \\
\bottomrule
\end{tabular}
\end{table}

The crossover family was established as the stronger latency oriented deployment choice on both platforms. On ESP32, Crossover-BiDir-BabyMamba-HAR completed all datasets with sub-300~ms latency, reaching 79.6~ms on Daphnet and 271.9~ms on Opportunity. The channel independent family remained heavier because it's recurrent backbone is executed independently for each sensor channel before pooling, with latency scaling strongly with channel count (reaching 49.6~s on Opportunity with 79 channels on Pico~2, and 8.4~s on ESP32).

\begin{table}[t]
\caption{Per dataset deployment results for Crossover-BiDir-BabyMamba-HAR.}
\label{tab:deploy_crossover}
\centering
\scriptsize
\setlength{\tabcolsep}{2pt}
\begin{tabular}{l|cc|cc}
\toprule
 & \multicolumn{2}{c|}{Pico~2} & \multicolumn{2}{c}{ESP32} \\
Dataset & Lat.~(ms)$\downarrow$ & Parity~(\%)$\uparrow$ & Lat.~(ms)$\downarrow$ & Parity~(\%)$\uparrow$ \\
\midrule
UCI-HAR & 507.0 & 99.99 & 147.6 & 99.70 \\
MotionSense & 500.6 & 99.91 & 154.6 & 99.30 \\
WISDM & 494.0 & 99.99 & 151.7 & 99.60 \\
PAMAP2 & 519.9 & 99.97 & 154.1 & 99.54 \\
Opportunity & 695.0 & 99.99 & 271.9 & 99.16 \\
UniMiB & 496.8 & 99.97 & 150.7 & 99.19 \\
Skoda & 396.8 & 99.99 & 125.4 & 97.20 \\
Daphnet & \textbf{245.2} & \textbf{100.00} & \textbf{79.6} & \textbf{99.92} \\
\bottomrule
\end{tabular}
\end{table}

Table~\ref{tab:deploy_crossover} presents the per dataset breakdown for Crossover-BiDir-BabyMamba-HAR. Parity remained above 99.9\% on Pico~2 and above 97\% on ESP32, with the slight ESP32 reduction attributable to INT8 projection dequantization. The ESP32 latency advantage (154.4~ms vs. 481.9~ms) reflects it's higher clock speed (240~MHz vs. 150~MHz).

\subsection{Comparison with Baseline Architectures}

To contextualize the BabyMamba deployment results against established HAR baselines, INT8 quantized and FP32 full precision deployment data for TinyHAR, TinierHAR, and DeepConvLSTM were obtained from a concurrent cross platform study conducted on the same Pico~2 and ESP32 hardware using the TFLite Micro runtime. Tables~\ref{tab:baseline_int8} and~\ref{tab:baseline_fp32} present the comparison. The BabyMamba variants achieved full 8/8 dataset coverage with above 99\% parity, whereas the INT8 TFLite baselines showed significantly degraded coverage and parity (TinyHAR: 7/8 coverage with 60.4\% parity on Pico~2, TinierHAR: 8/8 coverage with 54.2\% parity, DeepConvLSTM: 1/8 coverage with 25.5\% parity).

\begin{table}[t]
\caption{INT8 quantized deployment comparison across platforms.}
\label{tab:baseline_int8}
\centering
\scriptsize
\setlength{\tabcolsep}{2pt}
\begin{tabular}{l|ccc|ccc}
\toprule
 & \multicolumn{3}{c|}{Pico~2} & \multicolumn{3}{c}{ESP32} \\
Model & Runs$\uparrow$ & Lat.$\downarrow$ & Par.$\uparrow$ & Runs$\uparrow$ & Lat.$\downarrow$ & Par.$\uparrow$ \\
\midrule
Crossover-BiDir & \textbf{8/8} & 481.9 & \textbf{99.98} & \textbf{8/8} & \textbf{154.4} & \textbf{99.20} \\
CI-BabyMamba & \textbf{8/8} & 11762 & 99.94 & \textbf{8/8} & 2768 & 99.36 \\
\midrule
TinyHAR & 7/8 & 354.1 & 60.4 & 4/8 & 487.0 & 88.6 \\
TinierHAR & \textbf{8/8} & \textbf{229.3} & 54.2 & 6/8 & \textbf{204.5} & 90.8 \\
DeepConvLSTM & 1/8 & 768.8 & 25.5 & 0/8 & --- & --- \\
\bottomrule
\end{tabular}
\end{table}

\begin{table}[t]
\caption{FP32 full precision deployment comparison across platforms.}
\label{tab:baseline_fp32}
\centering
\scriptsize
\setlength{\tabcolsep}{2pt}
\begin{tabular}{l|ccc|ccc}
\toprule
 & \multicolumn{3}{c|}{Pico~2} & \multicolumn{3}{c}{ESP32} \\
Model & Runs$\uparrow$ & Lat.$\downarrow$ & Par.$\uparrow$ & Runs$\uparrow$ & Lat.$\downarrow$ & Par.$\uparrow$ \\
\midrule
Crossover-BiDir & \textbf{8/8} & 481.9 & 99.98 & \textbf{8/8} & 154.4 & 99.20 \\
CI-BabyMamba & \textbf{8/8} & 11762 & 99.94 & \textbf{8/8} & 2768 & 99.36 \\
\midrule
TinierHAR & 7/8 & \textbf{231.5} & \textbf{100.0} & 3/8 & \textbf{96.4} & \textbf{100.0} \\
TinyHAR & 5/8 & 486.1 & 89.6 & 3/8 & 352.9 & 88.1 \\
DeepConvLSTM & 1/8 & 2034.6 & 26.3 & 0/8 & --- & --- \\
\bottomrule
\end{tabular}
\end{table}

Several observations emerge from the baseline comparison. Under INT8 deployment, both BabyMamba families achieved full coverage on both platforms with above 99\% parity, whereas TFLite baseline parities degraded considerably (TinierHAR to 54.2\% on Pico~2, TinyHAR to 60.4\%). This parity gap is due to the handcrafted recurrent path retaining float32 state evolution, which avoids the cascading quantization errors that accumulate in graph compiled recurrent networks. Under FP32 deployment, TinierHAR recovered to 100\% parity on it's successful runs, confirming that the INT8 parity collapse was a quantization artifact. DeepConvLSTM remained non-deployable on both platforms regardless of precision, with it's 136K parameter footprint exceeding available tensor arena budgets.

\begin{figure}[t]
\centering
\includegraphics[width=\columnwidth]{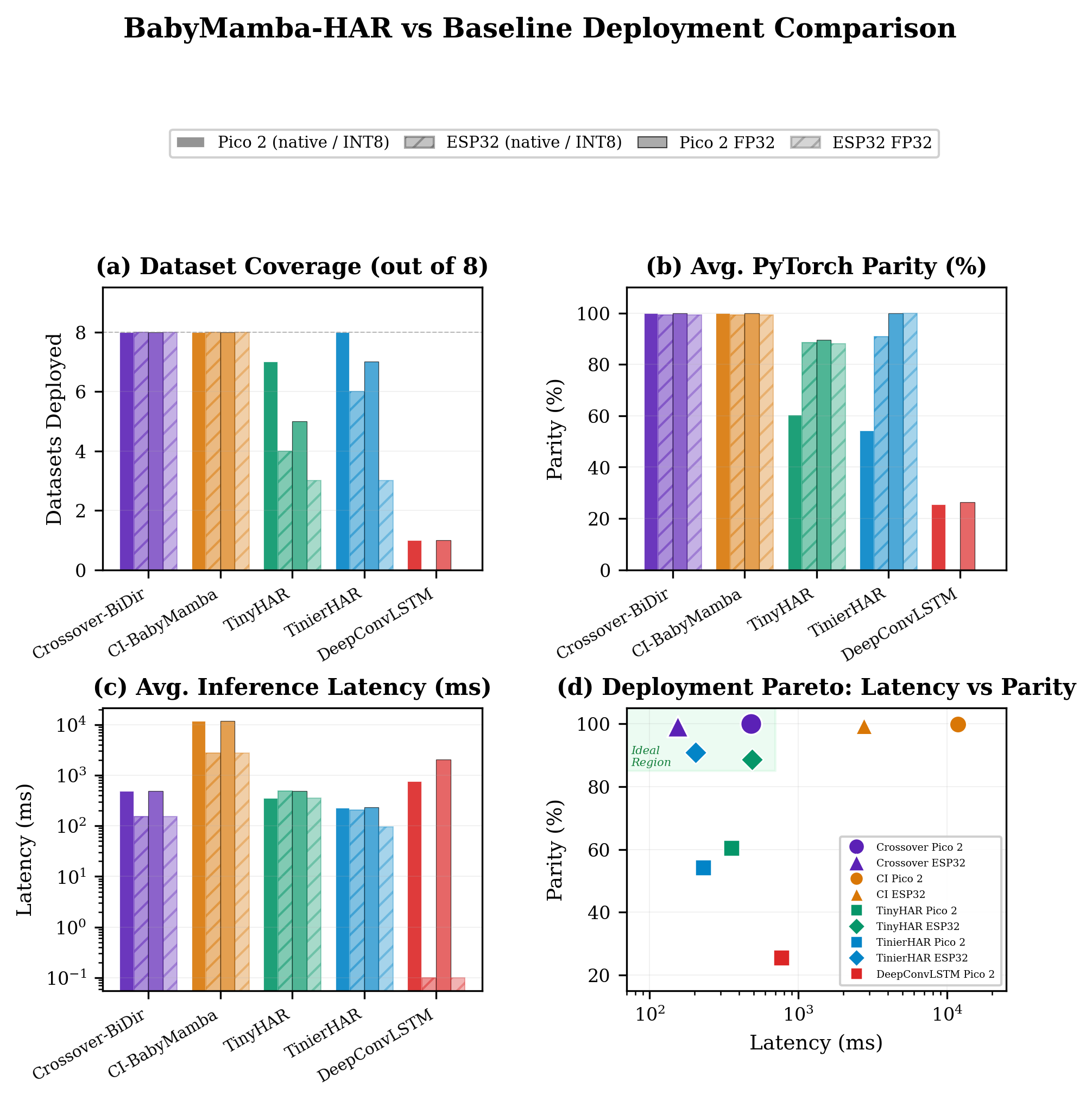}
\caption{Cross platform deployment comparison of BabyMamba families against TFLite baselines. (a)~Dataset coverage, (b)~average PyTorch parity, (c)~average inference latency, and (d)~deployment Pareto frontier.}
\label{fig:baseline_deploy}
\end{figure}

Figure~\ref{fig:baseline_deploy} visualizes the deployment landscape. The BabyMamba families occupy a distinct region of full coverage and near perfect parity, while TFLite baselines trade coverage and fidelity for lower per inference latency in their successful configurations. The mixed precision strategy (INT8 projections with float32 recurrence) was validated as effective for selective state space models on constrained microcontrollers, consistent with the optimization principles outlined in \cite{xu2025mambalite}.

\section{Discussion}

\subsection{Architecture Selection Guidelines}

The evaluation yields actionable deployment guidelines: (1)~\textbf{High channel count} ($C > 30$): Crossover-BiDir-BabyMamba-HAR achieves 11$\times$ fewer MACs than TinyHAR on Opportunity while matching F1-score, as backbone complexity is channel independent. (2)~\textbf{Heterogeneous sensors}: CI-BabyMamba-HAR isolates per-channel noise, preferred when sensors have independent noise profiles (PAMAP2). (3)~\textbf{Extreme constraints}: Crossover-BiDir-BabyMamba-HAR with $d_{\text{state}}=8$ achieves sub-25K parameters and $<$3M MACs. (4)~\textbf{Correlated features}: Early fusion strongly favored for pre-computed features (UCI-HAR: +9.49\% F1-score with Crossover-BiDir-BabyMamba-HAR).

\subsection{Efficiency and Scalability}

The channel independent complexity of Crossover-BiDir-BabyMamba-HAR is decisive for emerging high density sensor networks. Full body motion capture (50--100 channels) renders CI architectures impractical (222M MACs on Opportunity), while Crossover-BiDir variant maintains constant backbone cost regardless of $C$. The $O(N)$ sequence complexity of SSMs versus $O(N^2)$ attention further advantages BabyMamba-HAR for longer windows.

\subsection{Deployment Considerations}

The on-device results provide empirical deployment evidence beyond desktop profiling. The crossover family was identified as the clear latency oriented choice, reaching 154.4~ms average on ESP32, which is compatible with real time HAR at typical sampling rates (30 to 100~Hz). The channel independent family remained feasible on both platforms at higher cost, limiting it's on-device applicability to offline or batch processing scenarios.

\subsection{Key Design Insights}

The ablations reveal that bidirectional scanning contributes up to 8.42\% F1-score for complex temporal patterns, validating non-causal processing for windowed classification. The weight tied mechanism doubles receptive field without parameter overhead. Stem architecture choice is data dependent: C2 ablation across all 8 datasets confirms fusion superiority on 7 out of 8 benchmarks, with channel isolation beneficial only for explicitly noisy sensor configurations.

\subsection{Limitations}

Key limitations include: (1)~bidirectional design precludes true streaming inference, (2)~extreme class imbalance (Daphnet: $>$90\% negative) may benefit from focal loss, and (3)~evaluation is limited to inertial modalities. The on-device study was conducted with handcrafted C++ inference rather than through a standard graph compiled path, and the INT8 projection compression was applied only to the projection heavy matrices rather than to the full model. Future work includes streaming compatible variants, mixed stem routing, and quantization aware training for further latency reduction.

\section{Conclusion}

In this work, BabyMamba-HAR is presented as a framework of two novel lightweight selective state space architectures for efficient human activity recognition. CI-BabyMamba-HAR provides noise robustness through channel independent processing, while Crossover-BiDir-BabyMamba-HAR achieves channel count independent computational complexity through early fusion. Both architectures incorporate weight tied bidirectional scanning and context gated temporal attention pooling.

Through evaluation across eight diverse benchmarks, it is demonstrated that Crossover-BiDir-BabyMamba-HAR achieves 86.52\% average macro F1-score with $\sim$27K parameters and 2.21M average MACs, matching established baselines while offering 11$\times$ computational reduction on high channel datasets. On-device deployment on the Raspberry Pi Pico~2 and ESP32 was validated through a handcrafted recurrent C++ engine with mixed precision inference, following the principles of MambaLite-Micro \cite{xu2025mambalite}. Both families achieved full 8/8 dataset coverage on both platforms with above 99\% parity, while INT8 TFLite baselines showed degraded coverage and fidelity. Systematic ablations establish that bidirectionality (up to +8.42\% F1-score) and gated attention (up to +8.94\% F1-score) are important components, while hyperparameters show remarkable robustness in the lightweight regime.

These findings provide practical guidance for deploying selective state space models as TinyML backbones for human activity recognition, validated through empirical on-device measurements rather than through desktop simulation alone. Future work includes exploration of streaming compatible variants, mixed stems for grouped sensors, and quantization aware training for further latency reduction on additional MCU targets. Code, trained models, and deployment artifacts are available at: \url{https://github.com/WhiteMetagross/BabyMambaHAR}.

\end{document}